\documentclass[preprint,12pt]{elsarticle}

\usepackage{amssymb}
\usepackage{amsmath}
\usepackage{float}
\usepackage{array}
\usepackage{tabularx}
\usepackage{booktabs}
\usepackage{caption}
\usepackage{textcomp}

\journal{Advanced Engineering Informatics}

\begin{document}

\begin{frontmatter}



\title{Contrastive Augmented Transformer with Domain-specific Enhancement for Robust Multi-scenario Metal Surface Defect Detection\tnoteref{t1}}
\tnotetext[t1]{This research was supported by the Innovation Fund of Glasgow College, University of Electronic Science and Technology of China.}


\author[gc]{Yiyao Liu\fnref{equal}}
\ead{2023190505007@std.uestc.edu.cn}

\author[gc]{Wenxiao He\fnref{equal}}
\ead{2022190504031@std.uestc.edu.cn}

\author[iapcm]{Liyuan Ren\corref{cor1}}
\ead{ren_liyuan@iapcm.ac.cn}

\author[tsinghua]{Huan Wang}
\ead{wh.2021@tsinghua.org.cn}

\cortext[cor1]{Corresponding author.}
\fntext[equal]{These authors contributed equally.}

\affiliation[gc]{organization={Glasgow College, University of Electronic Science and Technology of China},
            city={Chengdu},
            postcode={611731},
            country={China}}
\affiliation[iapcm]{organization={Institute of Applied Physics and Computational Mathematics},
            city={Beijing},
            country={China}}
\affiliation[tsinghua]{organization={Department of Industrial Engineering, Tsinghua University},
            city={Beijing},
            country={China}}

\begin{abstract}
Metal surface defect detection is critical for maintaining product quality in industrial manufacturing. 
However, it faces significant challenges including limited annotated data, difficulty in identifying multi-scale subtle defects, and poor generalization across diverse scenarios. 
To address these issues, this paper proposes a novel Contrastive Augmented Transformer (CAT) framework for robust defect detection. 
CAT employs a hierarchical Swin Transformer backbone and redesigns the feature pyramid network to effectively fuse low-level textures with high-level semantics, enabling precise modeling of subtle and multi-scale defect patterns. To enhance robustness under real-world noise conditions, we propose a domain-specific "droplet" augmentation algorithm. 
Furthermore, we incorporate a hard negative mining strategy into the contrastive loss to strengthen the model's discrimination ability in ambiguous defect regions. 
Experimental results on the KolektorSDD2 dataset demonstrate that CAT achieves a pixel-level AUROC of 99.54\%, outperforming existing methods. 
Additionally, CAT exhibits superior generalization and robustness on three unseen datasets-KSDD1, MTD (tile defects), and MSDD (rail surface defects) demonstrating its potential for wide-scale industrial deployment.

\end{abstract}


\end{frontmatter}



\graphicspath{{new_template/}}

\section{Introduction}
Metal surface defect detection targets on identifying damaged surfaces that are different from normal instances. 
It stands as a cornerstone of intelligent manufacturing, 
serving as a critical barrier to ensure product reliability across industries such as automotive~\cite{Automotive,Automotive2,Automotive3}, aerospace~\cite{aero1,aero2}, and precision machinery~\cite{pm1,pm2}. 
From micro-scratches on engine components to corrosion spots on pipeline surfaces, timely identification of these anomalies prevents catastrophic failures, 
reduces maintenance costs, and safeguards production efficiency. In practical industrial production, the majority of manufactured products are normal, whereas defective samples are extremely scarce and exhibit high variability.
This presents a critical challenge for defect detection methods---achieving accurate detection under the constraint of limited anomalous data.

Traditional surface defect detection techniques--relying on handcrafted features like texture, statistical measures, and structural 
filters were used to detect irregularities. These methods do not require a large amount of data and their computation complexities can be 
reduced with pre-determined algorithm parameters. For example, Liu, et al.~\cite{Gabor} used Gabor wavelet features for textured surface defect detection, combined with an adaptive filter-selection scheme to reduce computational costs. 
These features were fed into a one-vs-all SVM classifier to distinguish defective from non-defective pixels.
Asha, et al.~\cite{Texture} published another study which applies multi-scale, multi-orientation Gabor wavelet transforms to patterned textiles. 
It divides the transformed image into periodic blocks and uses energy statistics of each block to identify defects.

However, traditional methods rely heavily on hand-crafted features and heuristic rules, 
so they generalize poorly across different materials, lighting conditions, viewpoints, and defect types. 
They also require extensive manual tuning and feature engineering, limiting scalability and often yielding lower detection accuracy.
In recent years, deep learning architectures have been widely adapted to task of surface defect detection. 
He et al.~\cite{resnet} combined Convolutional Neural network (CNN) layers and skip connection (ResNet), overperforming other similar models in downstream 
tasks including images classification, objects detection, semantic and instance segmentation. Some defect detection approaches such as 
MMR~\cite{mmr} also utilizes it as the backbone of extracting multi-scale feature maps.
Recently, Transformer-based networks have become the new standard, and variants optimized for industrial inspection have achieved remarkable progress.
For example, Zhu et al.~\cite{R3-SDDC} utilized an efficient Swin Transformer for robust surface defect classification on steel, while Hou and Zhang~\cite{R1-LRD}
developed a lightweight, real-time detection transformer specifically designed for surface defect systems. Gao et al.~\cite{swin2}. proposed a Swin Transformer variant with cross-window fusion to better model local details.

In comparison, unsupervised or self-supervised learning, which requires few or zero defective samples, is more suitable for this task. Recent advances have evolved beyond basic proxy tasks.
While early methods like CutPaste~\cite{CutPaste} and NSA~\cite{NSA} rely on synthetic patches, newer paradigms incorporate Masked Autoencoders (MAEs)---such as MAEDAY~\cite{MAEDAY}---and domain-adaptive contrastive approaches to learn more robust representations.
Notably, Huang et al.~\cite{R2-ACT} proposed an adaptive cross-transformer with contrastive learning, demonstrating the effectiveness of aligning multi-scale features for surface defect detection.
However, despite these advancements, existing methodologies often fall short when confronted with dynamic domain shifts, varying illumination, and pervasive real-world stochastic noise (e.g., oxidation spots) on metal surfaces.

Although approaches for general surface defect detection have become increasingly mature, translating lab-based algorithms into real-world industrial scenarios remains fraught with challenges. 
Compared with other defect detection tasks (e.g., fabric or glass), the high reflectivity and rigid textural characteristics of metal surfaces make distinguishing defects from backgrounds exceptionally difficult.
Existing methodologies often fall short when confronted with these intricate real-world conditions. Multi-scale anomalies ranging from coarse indentations to sub-millimeter cracks; dynamic domain shifts driven by fluctuating illumination and changing camera viewpoints; and pervasive stochastic noise (e.g., oxidation spots) severely test the models' generalization capabilities.
Consequently, models developed for specific conditions frequently exhibit brittle performance, failing to generalize across novel production lines.

To address these challenges, we propose the Contrastive Augmented Transformer (CAT), \textbf{a domain-specific framework that systematically integrates an asymmetric dual-branch architecture} 
for robust metal surface defect detection. Instead of inventing a completely new backbone, CAT leverages existing powerful feature extractors in a complementary manner:
\textbf{Swin Transformer to capture globally sensitive anomaly representations, alongside a frozen pre-trained ResNet-50 backbone to provide a stable reference for normal textures}.
To overcome the limitations of generic synthetic anomalies (like CutPaste), we introduce a physics-inspired "droplet" augmentation algorithm that accurately simulates real-world metal artifacts.
Furthermore, an enhanced MSF-FPN is coupled with a weighted multi-scale contrastive loss to effectively penalize hard negatives in ambiguous regions.

To evaluate CAT's performance, it is trained on KolektorSDD2~\cite{KOS2} and evaluated on multiple datasets.
CAT achieves \textbf{highly competitive performance} on the KolektorSDD2 dataset (pixel-level AUROC of 99.54\%) and exhibits \textbf{improved cross-domain robustness} 
across unseen datasets (KolektorSDD1, MTD, RSDD). 

The main contributions of this work are summarized as follows:
\begin{enumerate}
    \item \textbf{A domain-specific system-level framework}, Contrastive Augmented Transformer (CAT), is integrated. By combining a trainable Swin-T with a frozen ResNet-50 backbone, it leverages an asymmetric dual-branch design to balance anomaly sensitivity and normal-texture stability.
    \item To mitigate anomalous data scarcity and domain shifts, a physics-prior \textbf{Droplet Augmentation} strategy is proposed to simulate realistic metal-surface artifacts (e.g., oxidation spots).
    \item A \textbf{Weighted Multi-Scale Contrastive Loss} is formulated alongside an enhanced MSF-FPN. This integration dynamically reweights discriminative feature maps, focusing on hard negatives to enhance sensitivity to subtle defects.
    \item Extensive experiments demonstrate CAT's competitive accuracy on KSDD2 and its cross-domain adaptability on KSDD1, MTD, and RSDD, \textbf{highlighting its potential for practical deployment} in varied industrial scenarios.
\end{enumerate}

The remainder of this paper is structured as follows: Section II reviews related work on contrastive learning and metal surface defect detection; Section III details the architecture of the CAT framework; Section IV presents experimental results and ablation studies; Section V concludes with future directions.

\section{Related Works}\label{sec:rw}
Comprehensive surveys of AI-enabled industrial defect detection are available~\cite{add_3}.
\subsection{Self-supervised Contrastive Learning in Vision}
Self-supervised learning has emerged as a powerful paradigm for learning visual representations without extensive labeled data.
By contrasting positive and negative image pairs, models can capture discriminative features. For instance, early frameworks like MoCo~\cite{moco} and MoCoV2~\cite{mocov2} achieved great success in general-purpose visual tasks by utilizing momentum-based queues and basic augmentations.
Recently, domain-adaptive contrastive approaches have been developed specifically tailored to surface defects, utilizing adaptive cross-transformers to align normal and anomalous features~\cite{R2-ACT,recontrast}. 
However, when directly transferred to metal surfaces, their effectiveness is frequently constrained. Standard synthetic defect generation methods (such as CutPaste~\cite{CutPaste} or NSA~\cite{NSA}) and general-purpose augmentations (e.g., random cropping, basic geometric transformations) are insufficient to capture the specific physical characteristics of metal defects, like oxidation-induced micro-white spots or complex stochastic textures.
To bridge this gap, instead of relying on generic proxy tasks, we introduce a physics-inspired "droplet" augmentation strategy. This domain-specific approach provides more realistic positive and negative pairs for contrastive learning, thereby mitigating domain shifts and improving generalization in practical metal inspection scenarios.
\subsection{Industrial Defect Detection Approaches}
Traditional industrial defect detection methods rely on hand-crafted features and rule-based systems, such as Canny edge detection~\cite{Canny}.
However, these methods are sensitive to noise and require extensive parameter tuning. With the rise of deep learning, convolutional neural networks (CNNs), including U-Net~\cite{U-Net} and DeepLabV3+~\cite{DeepLabV3+}, have been widely applied to localize defects by learning pixel-wise representations. Yet, traditional CNN-based methods often struggle to effectively capture multi-scale features, which is crucial for detecting defects varying from large indentations to sub-millimeter cracks.

To overcome the receptive field limitations of CNNs, recent advancements have increasingly shifted toward hierarchical attention mechanisms and Transformer variants optimized for industrial inspection~\cite{R1-LRD, R3-SDDC, add_1}.
Simultaneously, Masked Autoencoder (MAE) based methods, such as MAEDAY~\cite{MAEDAY} and self-supervised image restoration tasks like FAIR~\cite{add_2}, have demonstrated strong capabilities in reconstructing normal background textures for anomaly localization.
Despite these advances, addressing the extreme data scarcity and subtle, multi-scale nature of metal surface anomalies remains a challenge. 
In our proposed CAT framework, we systematically integrate a dual-branch architecture combining a trainable Swin Transformer with an enhanced MSF-FPN. 
This system-level integration effectively balances global semantic context with high-resolution local details, offering a more robust alternative to standard CNN paradigms for metal defect detection.

\section{Methodology}\label{sec:me}
\begin{figure*}[!t]
    \centering
    \includegraphics[width=\textwidth]{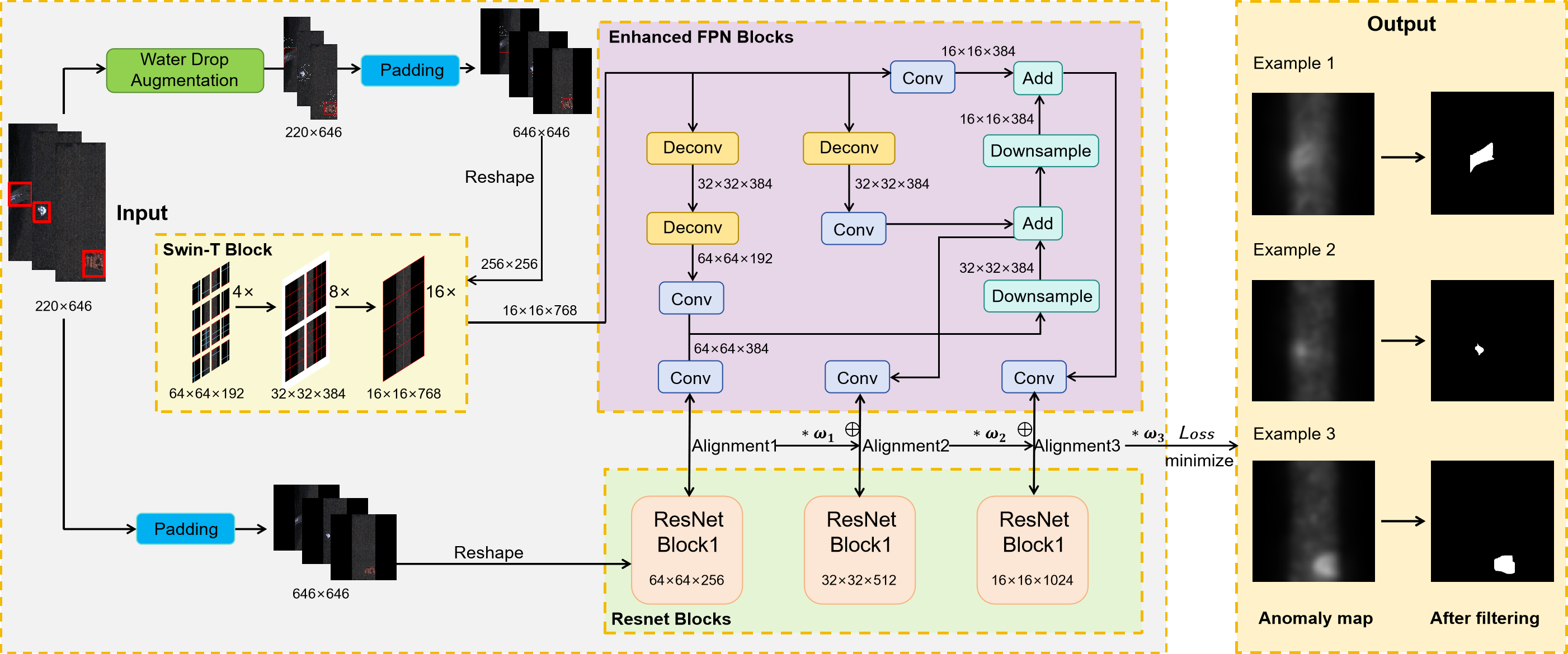}
    \caption{CAT's architecture. Images are first padded 
    into a fixed size. One copy of the padded image is fed into 
    a frozen, pre-trained encoder and another copy is ent into 
    droplet augmentation, swin transformer and enhanced FPN layer. The 
    multi-feature maps produced by two branches are then aligned helping 
    the model to learn features of the dataset.}
    \label{fig:arch1}
\end{figure*}

The Contrastive Augmented Transformer-based (CAT) framework is proposed to detect metal surface defects with superior accuracy and strong generalization ability. 
CAT is designed to overcome key challenges in metal surface defect detection including limited to defective data and variant domain shifts. 
It addresses these challenges through combining one core architecture and three strategies. The core architecture refers to asymmetric dual-branch backbone network (ADBN), which is developed by three strategies, including physics-inspired data augmentation, multi-scale feature fusion via MSF-FPN, and an adaptive loss with hard-negative weighting. 
As shown in Fig.~\ref{fig:arch1}, one branch generates synthetic defects through an augmentation algorithm, while the other processes the original defect-free images. Both branches extract multi-scale features, which are then compared to capture differences between normal and defective surfaces. In Section 3.1, the whole architecture of CAT is systematically introduced. 
Section 3.2 then introduces a new augmentation method named droplet augmentation, which simulates the physical properties of real anomalies. The following sections(3.3 and 3.4) describe the feature extraction modules and the adaptive loss function, which together improve the model’s robustness and discrimination ability.

\subsection{Asymmetric Dual-Branch Backbone Network}
The overall architecture of the proposed Contrastive Augmented Transformer (CAT) framework is constructed upon an asymmetric dual-branch backbone network (ADBN). 
The rationale for adopting such a design is to address two complementary objectives in defect detection: (i) to augment the limited and imbalanced training data by synthesizing realistic defects, and (ii) to preserve reliable representations of defect-free surfaces as a stable reference. 
By introducing asymmetry into the backbone design, CAT is capable of leveraging heterogeneous feature extractors tailored to these objectives, thereby enhancing its generalization ability under domain shifts and improving sensitivity to subtle anomalies. 
An overview of the framework is illustrated in Fig.~\ref{fig:arch1}.

The first branch focuses on anomaly-aware feature learning. 
Input images are first processed through the Droplet Augmentation module, which simulates physically plausible defects and increases the diversity of the training samples. 
The augmented images are then resized into a standardized square format to maintain structural consistency across the pipeline. 
A Swin-T backbone is subsequently employed to extract hierarchical features, benefiting from its transformer-based windowed self-attention for long-range context modeling. To further enrich the representation, 
a Multi-Scale Feature Pyramid Network (MSF-FPN) is applied, enabling both high-level semantic abstraction and fine-grained detail retention. This design ensures that the augmented branch captures defect-sensitive cues across scales.

In parallel, the second branch is dedicated to modeling defect-free surfaces. 
The original, unaltered images are fed into a pre-trained ResNet-50 backbone, which provides stable baseline features of normal textures and structures. 
This design allows CAT to decouple normal-surface representation from augmentation-driven variability, effectively suppressing the noise or artifacts that may arise from synthetic defect generation. 
The asymmetry between the transformer-based augmented branch and the CNN-based reference branch thus introduces complementary inductive biases, which together strengthen the robustness of the framework.

The two branches ultimately converge in a shared embedding space, where their feature representations are compared. 
A weighted contrastive loss is adopted to encourage alignment between features of similar regions while emphasizing discrepancies in defect-relevant areas. 
This adaptive loss function enables the network to focus on hard negative samples, i.e., defect-free regions that closely resemble anomalies, thereby improving its discrimination capability. 
The comparison results are decoded into an anomaly map and a filtered prediction map, which jointly constitute the detection outputs. Each component of this dual-branch design contributes synergistically to the overall framework, and the following subsections (3.2–3.4) elaborate on the data augmentation strategy, 
the multi-scale feature fusion mechanism, and the adaptive loss function in detail.
\begin{figure}[!t]
    \centering
    \includegraphics[width=0.5\textwidth]{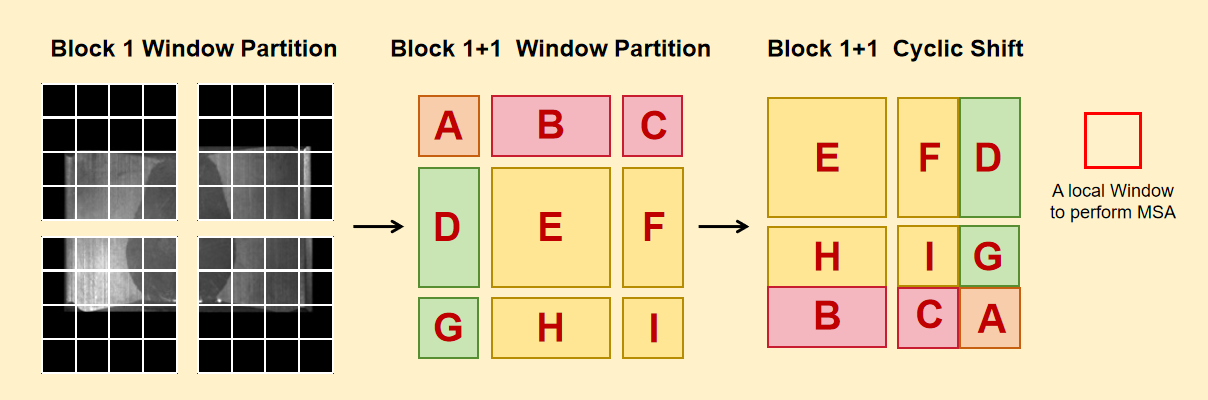}
    \caption{Structure of Swin Transformer block. One uses
Window-based Multi-Head Self-Attention (W-MSA) and
the other uses Shifted Window-based Multi-Head SelfAttention (SW-MSA).}
    \label{fig:swint}
\end{figure}

\subsection{"Droplet" Data Augmentation Strategy}
After the overall architecture is clear, CAT further relies on the data enhancement strategy inspired by physics.
Self-supervised models designed for defect detection require synthetic defect generation to learn patterns and representations 
of potential anomalies without defective data. 
However, many existing augmentation strategies are insufficient for this task. Approaches that rely on real defective
samples are fundamentally misaligned with self-supervised learning, as they assume availability of anomalies that are 
typically scarce in industrial environment. On the other hand, generic augmentation techniques---such as CutPaste~\cite{CutPaste} or NSA~\cite{NSA}
which introduce large domain shifts by manipulating or blending patches from samples, but they overlook the intrinsic 
physical characteristics of metal surface defects such as cracks, pits and uneven surfaces. As a result, 
the generated anomalies may not resemble the subtle, corrosion-like textures that commonly occur on metal surfaces. This potentially 
cause the overfit of artificial patterns rather than learning robust and generalizable representations. 

To overcome these limitations, we introduce a domain-specific augmentation method called droplet augmentation, inspired by the 
natural formation of metallic oxidation spots and corrosion-induced droplets. Such real-world defects typically appear as irregular,
blob-like regions with diffuse boundaries and scattered high-contrast points. By emulating these characteristics, 
the proposed method generates anomalies that are both realistic and diverse, helping CAT to capture 
features from normal textures and improving its robustness across different scenarios. 

Our proposed algorithm is designed to balance natural realism with controllability. 
It begins by randomly selecting a central point $(c_x, c_y)$ within the image to serve as the defect origin, 
ensuring diverse placement and mimicking the random occurrence of real anomalies. 
A regular polygon with $n_p$ vertices (typically 64–128 for sufficient resolution) and a base radius $R$ (sampled between 10 and half the image width) is then constructed
around the center.  
While this provides a symmetric foundation, realism is introduced by disordering the radius with 1D Perlin noise, a smooth, continuous function that induces wave-like variations. 
For each angle $\theta$ in an evenly spaced sequence, the new radius is computed as:
\[  r(\theta) = R + A_n \times N(\theta), \quad N(\theta) \in [-1, 1], \]
where $A_n$ (randomly selected between $\frac{R}{10}$ and $\frac{R}{2}$) controls the distortion magnitude. 
This process transforms the polygon into an irregular, fluid-edged region, avoiding the rigid boundaries and artificiality of purely geometric augmentations.

To further enhance realism, the interior of the region (denoted as $\Omega$) is sparsely populated with bright points simulating
oxidation spots or reflective pits. In our setting, up to $\frac{1}{16}$ of points inside $\Omega$ can be accepted with a low 
probability($\approx 0.01-0.1$). The accepted points are set to high intensity, forming clustered highlights without oversaturation. 
This stochastic sampling reflects the variability of natural corrosion patterns and prevents uniform or repetitive textures.

\begin{figure}[H]
    \centering
    \includegraphics[width=0.5\textwidth]{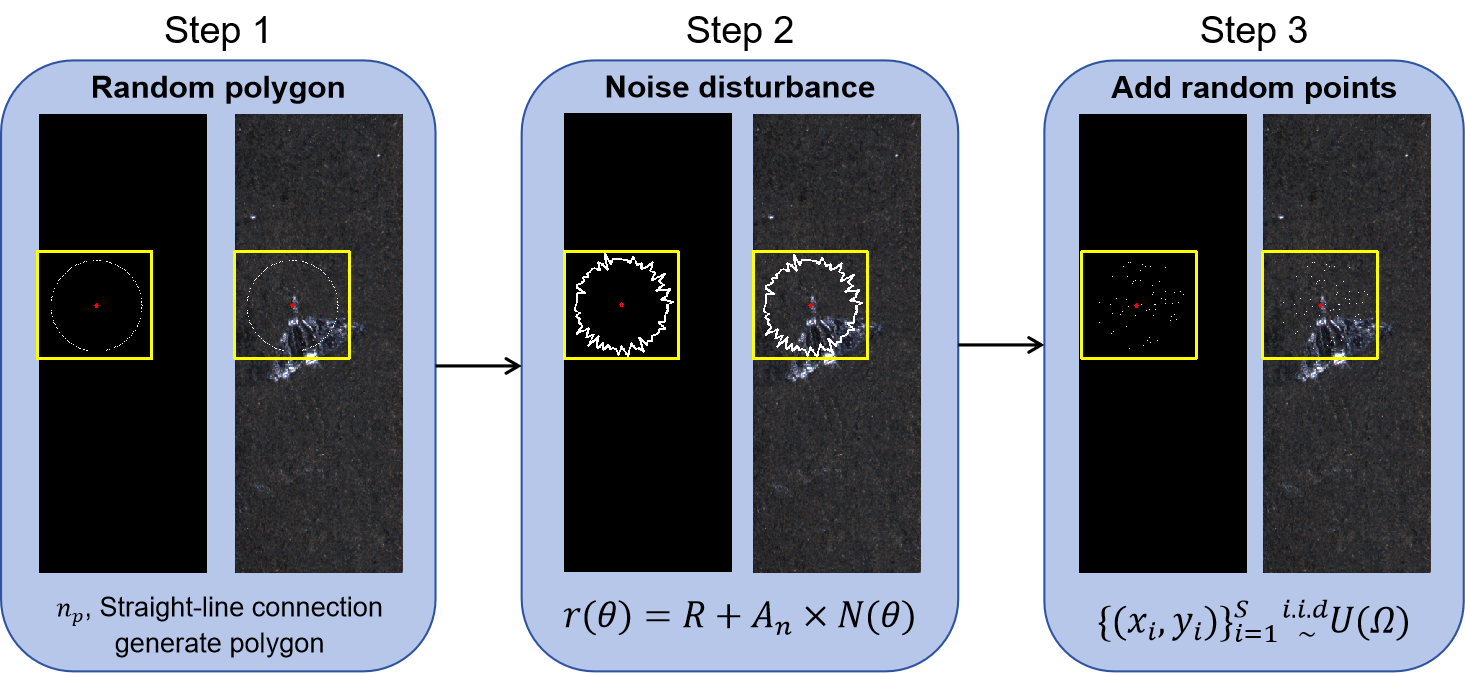}
    \caption{Process of droplet augmentation, it simulates physical damage of metal surfaces.}
    \label{fig:wd}
\end{figure}

The procedural flow is visualized in Fig.~\ref{fig:wd}, demonstrating how the method progressively builds from a 
simple polygon to textured anomaly that integrates seamlessly with the underlying metal surface. 
droplet augmentation creates a region of diverse shapes and does
not make drastic image alterations that differ substantially from real metal surface defects, assisting the CAT to extract and
learn better patterns of industrial defects during training. In section IV, we conducted a quantitative experiment on different anomalies 
generation methods and verify the superior performance of our proposed droplet augmentation algorithm.

\subsection{Multi-Scale Feature Fusion via MSF-FPN}
Conventional Feature Pyramid Networks (FPNs) suffer from inherent limitations when applied to defect detection. 
In particular, the vanilla top-down fusion pathway often leads to insufficient preservation of fine-grained textures from shallow layers, causing small or subtle defects to be overlooked. 
Moreover, the lack of bottom-up information flow results in semantic imbalance, where high-level features dominate but low-level details are underutilized. 
These challenges hinder the detection of multi-scale anomalies that require both fine spatial resolution and rich semantic context.

\begin{figure*}[!t]
    \centering
    \includegraphics[width=\textwidth]{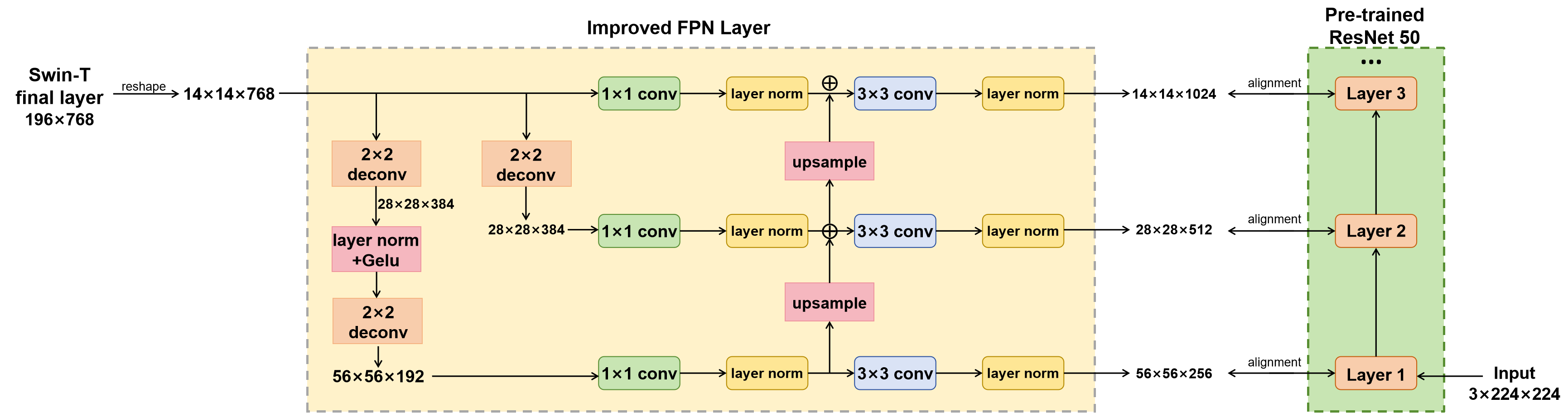}
    \caption{Multi-Scale Fusion FPN (MSF-FPN). A bi-directional fusion module combines top-down semantic upsampling ($2\times 2$ transposed conv) with bottom-up propagation of high-resolution textures. Features at each scale are merged via $1\times 1$ and $3\times 3$ convolutions, layer normalization and element-wise addition, producing multi-scale representations that preserve fine spatial details for micro-defect localization while retaining deep semantic context.}
    \label{fig:fpn}
\end{figure*}

To address these issues, the standard Feature Pyramid Network (FPN) is enhanced into a multi-scale fusion module (MSF-FPN, Fig.~\ref{fig:fpn}). 
Unlike the vanilla top-down pathway that solely propagates semantic features from higher to lower layers, our design introduces a bi-directional fusion strategy, where both top-down and bottom-up pathways jointly contribute to feature refinement.

Specifically, in the top-down pathway, higher-level semantic maps are upsampled using $2\times 2$ transposed convolutions, followed by bilinear interpolation and element-wise addition with lower-level maps. 
This enables the preservation of fine spatial details critical for localizing micro-defects. 
Conversely, in the bottom-up pathway, enriched low-level maps are progressively propagated upward, injecting high-resolution textures into deeper layers. 
This two-way information flow alleviates the imbalance between spatial precision and semantic abstraction that is common in conventional FPNs.

The fused representations simultaneously integrate low-level textures (e.g., scratches, edges) and high-level semantics (e.g., corrosion patterns, contextual cues), ensuring multi-scale coherence. 
Ablation studies confirm that this enhanced FPN significantly improves the detection of subtle, small-scale defects while maintaining robustness for large-area anomalies. 
Moreover, the outputs of the MSF-FPN are precisely aligned with the frozen branch, ensuring that both branches produce comparable multi-scale features for the subsequent weighted contrastive loss.

\subsection{Hard Negative Mining with Weighted Contrastive Loss Function}
Despite the improvements introduced by MSF-FPN, multi-scale feature fusion alone cannot fully resolve the imbalance between subtle and large-scale defects. 
Conventional loss functions typically treat all features equally, overlooking the fact that anomalies of different sizes contribute unequally to detection performance. 
This limitation often results in insufficient penalization of \textbf{hard negative samples}, where defect-free regions share similar textures with true anomalies. 
To address this challenge, and in line with our bi-directional fusion strategy, the weighted contrastive loss is designed that assigns adaptive importance to features at different scales. 
In this way, the loss function complements the pyramid architecture: while MSF-FPN enriches multi-scale representations, the adaptive weighting mechanism ensures that each scale’s contribution is appropriately emphasized during optimization.

To address this limitation, CAT incorporates a \textbf{weighted contrastive loss} that adaptively emphasizes 
hard negatives. The key idea is to assign larger penalties to negatives that are close to the anchor in feature space, 
thereby forcing the model to focus on the most confusing cases. This approach works synergistically with the multi-scale 
feature extraction module: the dual-branch backbone captures diverse and discriminative features, while the weighted loss ensures 
that these representations are actually shaped to separate subtle anomalies from normal variations. 

Formally, for an anchor sample $\alpha$, its positive counterpart $p$ (augmented from the same image), and a set of negatives 
$N$, we first define cosine similarity between feature vectors $f(x)$ and $f(y)$ as:
\begin{equation}
    s(x,y)=\cos(f(x),f(y))=\frac{f(x)\cdot f(y)}{||f(x)||||f(y)||}
\end{equation} 
with cosine distance $d(x,y)=1-s(x,y)$. The weighted contrastive loss is then
\begin{equation}
    \mathcal{L}=\mathbb{E}[d(a,p)+\sum_{n\in N} \omega_n \cdot d(a,n)]
\end{equation}
To emphasize negatives with higher similarity (hard negatives), we compute normalized similarity-based weights via a softmax:
\begin{equation}
    \omega_n = \frac{\exp(\tau\, s(a,n))}{\sum_{n'\in N} \exp(\tau\, s(a,n'))},
\end{equation}
where \(\tau>0\) is a temperature parameter controlling the concentration of weights.
This formulation ensures the negatives lying closer to anchor---those most likely to be confused with defects---dominate the optimization. 

The adaptive weighting scheme is particularly impactful in detecting metal surface defects, 
where a core challenge lies in distinguishing subtle scratches from normal textural variations.
CAT's loss function directly counteracts this by magnifying the penalty for these ambiguous negatives, forcing 
the model to learn fine-grained feature representations, leading to a significant reduction in both false alarms and missed detections. 
Thus, the method successfully adapts a generic contrastive learning framework to meet the domain-specific precision required in 
industrial settings, ensuring greater robustness in practice. 
%

\section{Experiment}\label{sec:exp} 
\subsection{Datasets \& Evaluation Metrics}
\begin{figure*}[!t]
    \centering
    \includegraphics[width=0.6\textwidth]{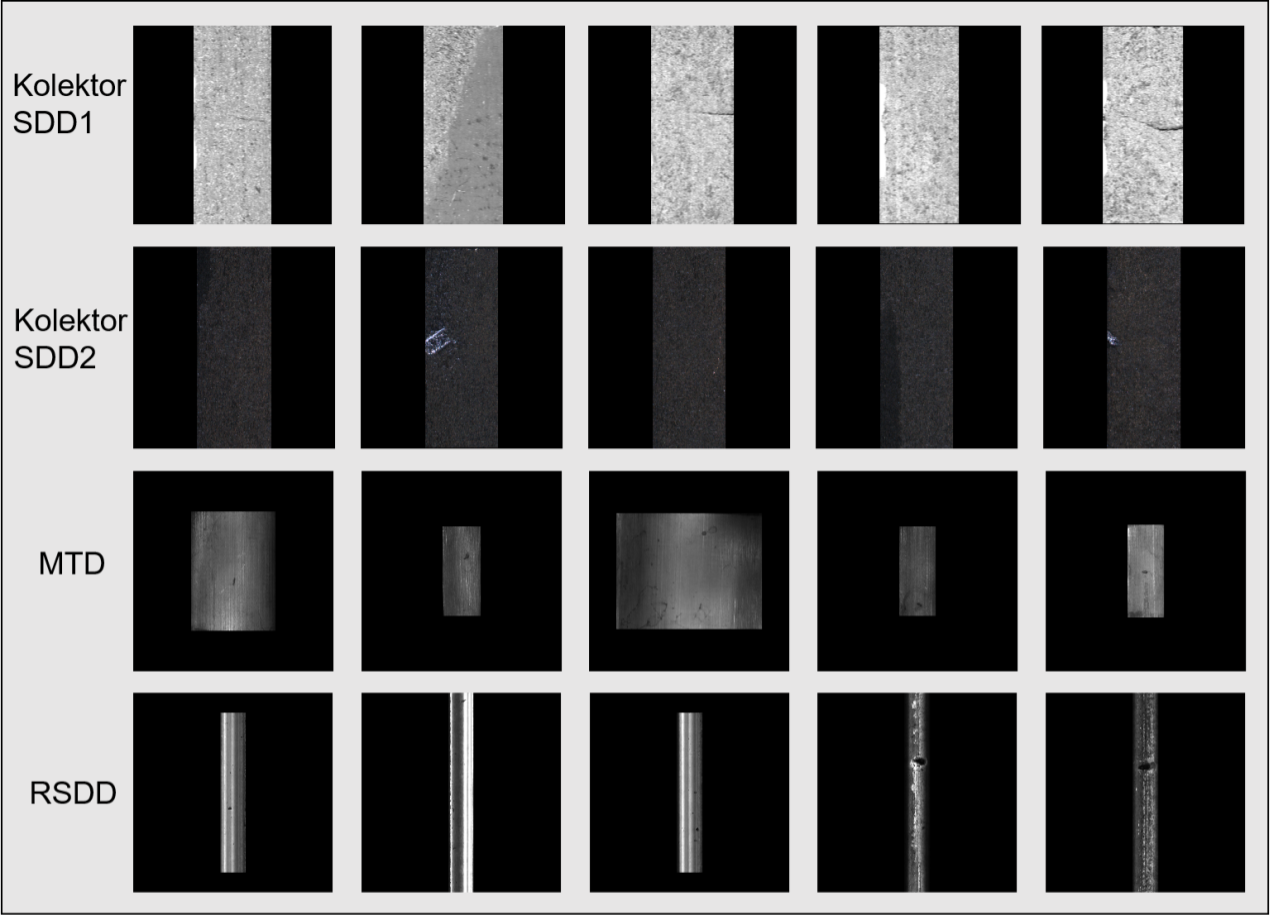}
    \caption{Representative examples of defect samples from four datasets (KolektorSDD1, KolektorSDD2, MTD and RSDD).} 
    \label{fig:datasets}
\end{figure*}

To validate the robustness and generalization of the Contrastive Augmented Transformer (CAT) framework for metal surface defect detection, 
the experiments are conducted on four benchmark datasets covering diverse industrial scenarios, 
with a primary focus on the KolektorSDD2 (KSDD2) dataset (as the core training and evaluation dataset) and supplementary verification on three unseen datasets. 
Representative examples of defect samples from each dataset are shown in Fig.~\ref{fig:datasets}

\begin{itemize}
    \item KolektorSDD2 (KSDD2, Core Dataset): As the primary training and evaluation dataset, it focuses on metal component surfaces and is optimized for the small-sample industrial scenario. It contains 2,331 training images (exclusively normal samples, consistent with real industrial data scarcity of defective samples) and 1,005 evaluation images (894 normal, 111 defective), but the training images only contains the normal samples. Defect categories include surface scratches (width 2–5 px), linear cracks (length 10–50 px), and irregular corrosion marks (area 20–100 $px^2$)—all typical subtle defects in automotive/aerospace component manufacturing.
    \item KolektorSDD1 (KSDD1, Supplementary Dataset): The earlier version of KSDD2, with 500 total images (380 normal, 120 defective), wihch has detailed pixel-level labeling. It is used to verify CAT's backward compatibility with low-annotation datasets and cross-version generalization.
    \item MTD (Magnetic Tile Defects, Cross-Material Dataset): A non-metal surface dataset (magnetic tiles) with 1,200 images (900 normal, 300 defective) to test CAT's cross-material adaptability. It features complex background textures (e.g., grid-like tile patterns) and ultra-small anomalies (e.g., pinholes with diameter $<$3 px, hairline cracks $<$1 px wide)—challenging the model's ability to distinguish defects from texture noise.
    \item RSDD (Rail Surface Defect Detection, Cross-Scenario Dataset): Focuses on large-scale rail surfaces, with 800 high-resolution images (600 normal, 200 defective) and variable illumination (e.g., overexposure in sunlight, underexposure in tunnels). Defects are long, thin scratches (length 50-200 px, width 1-3 px), simulating real-world dynamic domain shifts in railway inspection and verifying CAT's robustness to environmental changes.
\end{itemize}

The datasets are utilized in a complementary manner: the KolektorSDD2 dataset serves as the sole source for model training, 
while the KolektorSDD1, MTD, and RSDD datasets are reserved exclusively for evaluation, enabling a rigorous assessment of generalization across diverse surface textures and defect morphologies.

To comprehensively evaluate the model's performance, we adopt both image-level and pixel-level metrics, which provide complementary perspectives:
\begin{itemize}
    \item Image-level AUROC (AUC): Measures the ability to classify normal vs. defective images.
    \item Pixel-level AUROC: Evaluates the quality of anomaly localization at the pixel level.
    \item Per-region Overlap (IoU): Quantifies the spatial alignment between predicted and ground-truth defect regions.
    \item $\text{AP}_\text{det}$ (Detection Average Precision): Reflects the detection precision-recall tradeoff.
    \item $\text{AP}_\text{loc}$ (Localization Average Precision): Emphasizes defect boundary quality and localization accuracy.
\end{itemize}
\subsection{Implementation Details}
This section describes the hardware, software, and training configurations required to reproduce our experiments. 
All experiments were implemented using PyTorch 2.8.0 with CUDA 12.8. We utilized a single NVIDIA A800 80GB PCIe GPU. 
The random seed was set to 54 for all experiments to ensure reproducibility.

The model's backbone is ResNet50, which was initialized with pretrained weights. 
We used the Adam optimizer with $\beta_1=0.9$ and $\beta_2=0.95$. 
The initial learning rate was 0.0005, and the batch size was 64. The model was trained for 200 epochs.

All images were resized to $\mathbf{256\times 256}$ pixels. Standard augmentations included Random Horizontal Flip with a probability of 0.5, applied only during training. 
Images were normalized using a mean of $[0.485, 0.456, 0.406]$ and a standard deviation of $[0.229, 0.224, 0.225]$. 
Additionally, our unique droplet augmentation was applied to 100\% of the training images , with a random number of points ranging from 25 to 128 and a random radius from 10 up to the image width.

\subsection{Ablation on Model Components}
\label{subsec:ablation}

To evaluate the contribution of each key component within the CAT framework, we conducted a series of ablation studies on the KolektorSDD2 dataset. Our analysis is divided into two parts: a component-only analysis to assess the individual impact of each module and a removal analysis to understand its indispensable role within the full framework.

\subsubsection{Component-only Analysis}
\label{subsubsec:component_only}

Component independence analysis was conducted on the KolektorSDD2 dataset to verify the individual contributions of each core module in CAT. As illustrated in Table~\ref{tab:component_only},
the experiment adopted an incremental verification strategy, starting from a baseline model and sequentially adding single components to evaluate performance changes. The baseline model (without A/B/C/D) uses Vision Transformer (ViT) as the backbone network, combined with a standard FPN and traditional $\ell_2$ reconstruction loss, achieving an Image-level AUROC of 89.7\%, Full Pixel-level AUROC of 99.47\%, Per-region Overlap of 95.10\%, detection accuracy $\text{AP}_{\text{det}}=0.77$, and localization accuracy $\text{AP}_{\text{loc}}=0.37$, reflecting the limitations of traditional ViT architecture in locating subtle defects on metal surfaces.

Replacing the backbone with Swin-T (A only) boosts Image-level AUROC to $92.00\%$ (an increase of $+2.30$ percentage points), 
Full Pixel-level AUROC to $99.48\%$ ($+0.01$ pp), Per-region Overlap to $95.30\%$ ($+0.20$ pp), $\text{AP}{\text{det}}$ to $0.80$ ($+0.03$), and $\text{AP}{\text{loc}}$ to $0.39$ ($+0.02$). 
This shows Swin-T’s windowed self-attention improves global feature aggregation and yields notable gains in image-level discrimination and region overlap, with modest localization benefits.

Introducing MSF-FPN (B only) raises Image-level AUROC to $91.30\%$ ($+1.60$ pp), Full Pixel-level AUROC to $99.53\%$ ($+0.06$ pp), 
Per-region Overlap to $95.60\%$ ($+0.50$ pp), $\text{AP}{\text{det}}$ to $0.79$ ($+0.02$), and $\text{AP}{\text{loc}}$ to $0.38$ ($+0.01$). 
These results confirm MSF-FPN’s role in fusing low-level textures with high-level semantics: it particularly improves region-level localization and pixel discrimination while only modestly affecting image-level classification.

Applying domain-specific Droplet augmentation (C only) increases Image-level AUROC to $91.70\%$ ($+2.00$ pp), Full Pixel-level AUROC to $99.51\%$ ($+0.04$ pp), leaves Per-region Overlap unchanged at $95.10\%$, and improves $\text{AP}{\text{det}}$ to $0.79$ ($+0.02$) and $\text{AP}{\text{loc}}$ to $0.40$ ($+0.03$). 
This indicates that simulating realistic defect morphologies (oxide spots, corrosion pits) enhances robustness to complex edges and irregular defects, producing detection and localization gains comparable to those from MSF-FPN.

Replacing the $\ell_2$ reconstruction loss with the weighted contrastive loss (D only) yields Image-level AUROC $=91.80\%$ ($+2.10$ pp), Full Pixel-level AUROC $=99.50\%$ ($+0.03$ pp), Per-region Overlap $=95.10\%$, $\text{AP}{\text{det}}=0.80$ ($+0.03$), and $\text{AP}{\text{loc}}=0.39$ ($+0.02$). 
This demonstrates that weighted contrastive learning—via hard negative mining and higher weighting of negatives near anchors—strengthens feature discrimination in ambiguous/blurred defect regions and supports improved detection and localization.

\begin{table*}[ht]
    \captionsetup{justification=centering, labelsep=newline}
    \centering
    \caption{Component-only Analysis Results on KolektorSDD2 Dataset, showing the individual performance gains of each CAT component.}
    \label{tab:component_only}
    \scriptsize
    \begin{tabularx}{\textwidth}{l *{5}{>{\centering\arraybackslash}X}}
        \toprule
        Configuration & Image-level AUROC & Full Pixel-level AUROC & Per-region Overlap & AP\textsubscript{det} & AP\textsubscript{loc} \\
        \midrule
        Baseline (no A/B/C/D) & 89.70\% & 99.47\% & 95.10\% & 0.77 & 0.37 \\
        A only (Swin-T) & \textbf{92.00\%} & 99.48 \% & 95.30\% & \textbf{0.80} & 0.39 \\
        B only (MSF-FPN) & 91.3\% & \textbf{99.53\%} & \textbf{95.60\%} & 0.79 & 0.38 \\
        C only (Droplet Augmentation) & 91.70\% & 99.51\% & 95.10\% & 0.79 & \textbf{0.40} \\
        D only (Weighted Contrastive Loss) & 91.80\% & 99.50\% & 95.10\% & \textbf{0.80} & 0.39 \\
        \bottomrule
    \end{tabularx}
\end{table*}

\subsubsection{Removal Analysis}
\label{subsubsec:removal}

To verify the indispensability of each core component in the CAT framework, performance degradation was quantified by sequentially removing single components from the complete model (A+B+C+D). 
As shown in Table~\ref{tab:removal}, the removal of any module leads to performance degradation, confirming that each component is indispensable for the high performance of CAT. The module whose removal causes the largest drop identifies its pivotal role in the framework.

The removal of the Swin-T backbone (-A) causes the most severe and widespread performance degradation. The image-level AUROC falls from 93.80\% to 91.30\% ($\Delta$AUROC=2.50\%), while AP\textsubscript{det} decreases from 0.824 to 0.795 ($\Delta$AP\textsubscript{det}=0.029), and AP\textsubscript{loc} drops from 0.49 to 0.40 ($\Delta$AP\textsubscript{det}=0.09). 
Despite pixel-level AUROC drops by 0.05\% (99.54\% $\rightarrow$ 99.49\%), the large declines in high-level classification and detection precision highlight the critical importance of Swin-T for capturing long-range dependencies and global context. 
Without this backbone, the model loses significant semantic discriminative power, directly impacting both classification and localization.

The removal of the Weighted Contrastive Loss (-D) also results in noticeable degradation, especially in localization-related metrics. AP\textsubscript{loc} decreases from 0.490 to 0.450 ($\Delta$AP\textsubscript{loc}=0.040), 
and per-region overlap drops from 95.80\% to 93.90\% ($\Delta$PRO=1.90\%). 
At the same time, AP\textsubscript{det} falls from 0.824 to 0.820 ($\Delta=0.004$). 
These results indicate that the Weighted Contrastive Loss plays a central role in mining hard negatives and improving boundary-level discrimination. Without it, the model struggles to refine localization consistency, particularly when defect boundaries blend with complex backgrounds.

The removal of Droplet Augmentation (-C) produces a moderate decline, especially in generalization-oriented metrics. 
AP\textsubscript{loc} decreases to 0.470 ($\Delta$AP\textsubscript{loc}=0.020), and AP\textsubscript{det} slightly falls to 0.823 ($\Delta$AP\textsubscript{det}=0.001), while image-level AUROC decreases to 93.56\% ($\Delta$AUROC=0.24\%). 
This shows that although Droplet Augmentation does not dominate in pixel-level accuracy, it enhances the model’s robustness against irregular and morphology-specific defect patterns (e.g., corrosion-like or spot-like defects). Its absence weakens the model’s generalization capability, resulting in less stable localization during evaluation.

The removal of MSF-FPN (-B) leads to the least degradation among the four components, but its effect is still non-negligible. Image-level AUROC decreases slightly to 93.60\% ($\Delta$AUROC=0.20\%), pixel-level AUROC decreases to 99.53\% ($\Delta$AUROC=0.01\%), and AP\textsubscript{loc} drops to 0.470 ($\Delta$AP\textsubscript{loc}=0.020). This suggests that MSF-FPN mainly provides multi-scale feature refinement, boosting robustness in detection and localization, though its contribution is more subtle compared to Swin-T and the weighted loss.

In summary, the results of the removal analysis demonstrate that Swin-T (A) is the most critical component, 
providing indispensable global context for both classification and precise localization. The Weighted Contrastive Loss (D) and Droplet Augmentation (C) strengthen robustness and fine-grained discrimination, ensuring reliable localization across complex defect morphologies. MSF-FPN (B) serves as a complementary enhancer by improving feature fusion across scales. 
Only when all components are combined can CAT achieve its best performance of 93.80\% image-level AUROC, 99.54\% pixel-level AUROC, and 0.490 AP\textsubscript{loc} on the KolektorSDD2 dataset.
\begin{table*}[ht]
    \captionsetup{justification=centering, labelsep=newline}
    \centering
    \caption{Removal Analysis Results on KolektorSDD2 Dataset, illustrating the performance degradation when each component is removed from the full CAT framework.}
    \label{tab:removal}
    \scriptsize
    \begin{tabularx}{\textwidth}{l *{5}{>{\centering\arraybackslash}X}}
        \toprule
        Configuration & Image-level AUROC & Full Pixel-level AUROC & Per-region Overlap & AP\textsubscript{det} & AP\textsubscript{loc} \\
        \midrule
        Full (A+B+C+D) & \textbf{93.80\%} & \textbf{99.54\%} & \textbf{95.80\%} & \textbf{0.824} & \textbf{0.49} \\
        -A (no Swin-T) & 91.30\% & 99.49\% & 95.10\% & 0.795 & 0.40 \\
        -B (no MSF-FPN) & 93.60\% & 99.53\% & 95.70\% & 0.815 & 0.47 \\
        -C (no Droplet Aug.) & 93.56\% & 99.54\% & 95.70\% & 0.823 & 0.47 \\
        -D (no Weighted Loss) & \textbf{93.80\%} & 99.53\% & 93.90\% & 0.820 & 0.45 \\
        \bottomrule
    \end{tabularx}
\end{table*}

In conclusion, the ablation studies confirm that each component of the CAT framework plays a crucial role in achieving the best results. 
The MSF-FPN is the most indispensable module for accurate localization. The droplet augmentation and weighted loss improve the model’s robustness and stability, particularly in handling the challenging KolektorSDD2 dataset. 
The Swin-T backbone enhances the model's ability to capture global representations. By synergistically combining these components, the full CAT model achieves the best trade-off across all evaluation metrics.

\subsection{Effect of Water-Drop Augmentation}
\begin{figure*}[!t]
    \centering
    \includegraphics[width=0.6\textwidth]{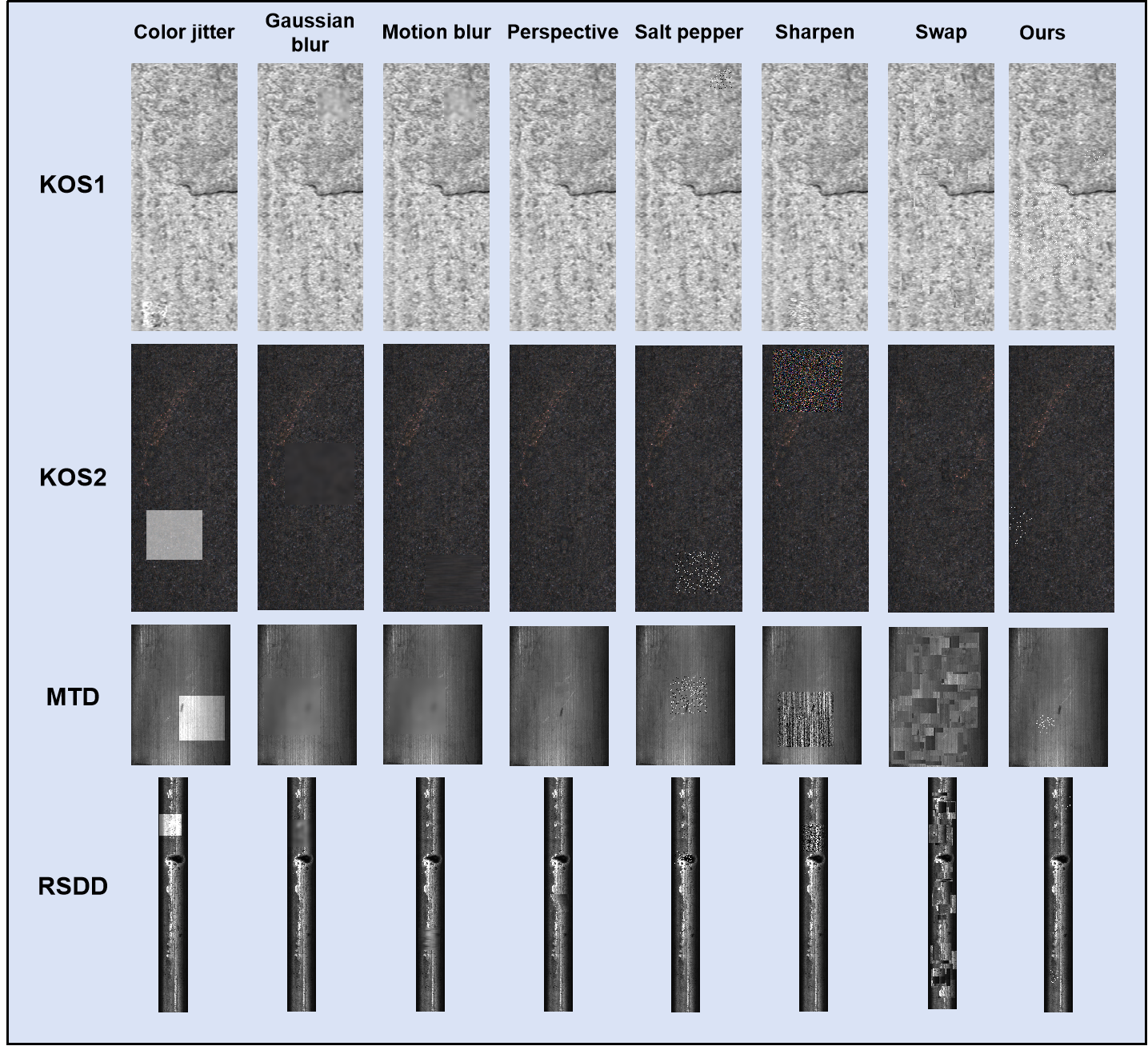}
    \caption{Example results of eight synthetic defect‐generation methods applied to one sample image from each of the four surface‐inspection datasets (rows: KSDD1, KSDD2, MTD, RSDD). From left to right, columns show (1) color jitter, (2) Gaussian blur, (3) motion blur, (4) perspective transform, (5) salt‐and‐pepper noise, (6) sharpening, (7) patch swapping, and (8) our proposed approach.}
    \label{fig:aug}
\end{figure*}
To assess the effectiveness of our proposed domain-specific droplet augmentation, we conducted a dedicated study on its impact. 
We compared the performance of our model using droplet augmentation against a baseline with no augmentation and a model with traditional data augmentation methods (including random patch swapping, gaussian blur, and color jitter). 
The results are shown in Table~\ref{tab:defect_comparison} and Fig.~\ref{fig:aug}.
\begin{table}[h]
    \captionsetup{justification=centering, labelsep=newline}

    \centering
    \caption{Comparison of Synthetic Defect Generation Methods on KolektorSDD2 Dataset}
    \label{tab:defect_comparison}
    \begin{tabular}{l c c}
        \toprule
        Method          & {Mean SSIM} & {Mean PSNR (dB)} \\
        \midrule
        Perspective transform & 0.378  & 18.89 \\
        Sharpening      & 0.350        & 22.35\\ 
        Motion blur     & 0.422        & 25.16 \\ 
        Color jitter    & 0.378        & 19.69             \\
        Patch swapping  & 0.401        & 24.87             \\
        Gaussian blur   & 0.390        & 23.72             \\
        \textbf{Water-drop} & \textbf{0.436} & \textbf{25.27} \\
        \bottomrule
    \end{tabular}
    \begin{flushleft}
        \small Note: Higher values of SSIM and PSNR indicate greater similarity to real defects. 
    \end{flushleft}
    \label{tab:augcmp}
\end{table}

As our droplet augmentation is designed to generate synthesized defects that closely resemble real-world defects, we first evaluate its effectiveness from a quantitative perspective. 
We compared the synthetic defects generated by our method with those from other common augmentation techniques using image quality metrics, namely Mean SSIM and Mean PSNR. 
Higher values for both metrics indicate greater similarity to real-world defects.

As shown in Figure~\ref{fig:aug}, the qualitative results highlight the superior performance of our droplet augmentation method in detecting complex defects. 
The model without augmentation often fails to distinguish subtle defects from the background, resulting in a low confidence score or completely missing the defect. 
In contrast, the model trained with droplet augmentation is more robust, accurately locating and segmenting subtle defects like fine scratches and cracks. 
This is because the synthesized water droplets closely resemble certain types of real-world defects, such as corrosion marks or small blemishes, effectively enhancing the model's ability to identify similar patterns in genuine defective samples.

\subsection{Comparison with Existing Methods}
\label{subsec:comparison}
Table~\ref{tab:comparison} presents a quantitative comparison of CAT with four representative 
unsupervised or semi-supervised defect-detection methods 
(SuperSimpleNet~\cite{ssn1,ssn2}, ReContrast~\cite{recontrast}, 
MMR~\cite{mmr}, and DFR~\cite{dfr}) 
as well as three general-purpose semantic segmentation networks 
(PSPNet~\cite{pspnet}, CCNet~\cite{ccnet}, and DeepLabV3+MobileNet~\cite{DeepLabV3,mobilenet}) 
on the KolektorSDD2 benchmark.  
These baselines span diverse methodological paradigms—reconstruction-based, 
contrastive-learning, feature-fusion, and supervised segmentation—providing a 
comprehensive context for evaluating CAT’s performance in industrial defect detection.

As shown in Table~\ref{tab:comparison}, 
CAT achieves the highest pixel-level discrimination 
(Pixel-level AUROC = 99.54\%) 
and competitive image-level classification 
(Image-level AUROC = 93.80\%, tied with DFR).  
It also attains strong region-level and detection metrics 
(Per-region Overlap = 95.80\%, 
$\mathrm{AP}_{\mathrm{det}} = 0.82$, 
$\mathrm{AP}_{\mathrm{loc}} = 0.49$).  
Compared with the strongest baselines, 
CAT improves pixel-level AUROC by 0.39 percentage points over MMR (99.45\%) 
and increases localization precision by 0.03 over ReContrast 
($\mathrm{AP}_{\mathrm{loc}} = 0.46$).

SuperSimpleNet, as a lightweight reconstruction-based baseline, 
achieves Image-level AUROC = 87.60\% and Pixel-level AUROC = 99.10\%.  
Although competitive in detection accuracy, 
it falls short of CAT in both classification and localization.  
ReContrast (Image AUROC = 90.86\%, Pixel AUROC = 99.11\%) 
and MMR (Image AUROC = 92.29\%, Pixel AUROC = 99.45\%) 
exhibit complementary strengths:  
ReContrast provides higher detection accuracy but lower region overlap, 
while MMR demonstrates better pixel discrimination yet weaker 
detection and localization scores.  
DFR performs competitively (Image AUROC = 93.80\%, Pixel AUROC = 99.15\%) 
and achieves the highest region overlap among baselines (96.30\%), 
but it remains inferior to CAT on detection and localization metrics.

The three general-purpose segmentation models—PSPNet, CCNet, 
and DeepLabV3+MobileNet—show clearly inferior performance 
on pixel-level and region-level indicators, 
despite achieving reasonable image-level AUROC values 
(PSPNet: 91.88\%, CCNet: 90.67\%, MobileNet: 89.37\%).  
In particular, PSPNet obtains a Pixel-level AUROC of 79.16\% 
and Per-region Overlap of only 5.52\%.  
These results demonstrate that standard segmentation networks 
trained on limited data are unable to capture the sparse, subtle, 
and multi-scale defects typical of industrial scenarios, 
and therefore fail to achieve accurate defect localization.

\begin{table*}[ht]
    \captionsetup{justification=centering, labelsep=newline}

    \centering
    \caption{Comparison with Existing Methods on KolektorSDD2 Dataset}
    \label{tab:comparison}
    \scriptsize
    \begin{tabular}{lccccccc}
        \toprule
        Method & Image-level AUROC & Pixel-level AUROC & Per-region Overlap & AP\textsubscript{det} &  AP\textsubscript{loc}  \\
        \midrule
        SuperSimpleNet & 87.60\% & 99.10\% & 94.49\% & 0.82 &  0.41  \\
        ReContrast & 90.86\% & 99.11\% & 91.95\% & \textbf{0.82} &  0.46  \\
        MMR & 92.29\% & 99.45\% & 95.50\% & 0.74 &  0.36  \\
        DFR & \textbf{93.80\%} & 99.15\% & \textbf{96.30\%} & 0.76 & 0.37 \\
        PSPNet & 91.88\% & 79.16\% & 5.52\% & 0.80 & 0.42 \\
        CCNet & 90.67\% & 88.80\% & 6.59\% & 0.57 & 0.36 \\
        MobileNet & 89.37\% & 73.79\% & 4.49\% & 0.79 & 0.30 \\
        CAT & \textbf{93.80\%} & \textbf{99.54\%} & 95.80\% & \textbf{0.82} &  \textbf{0.49}  \\
        \bottomrule
    \end{tabular}
\end{table*}

In summary, CAT consistently delivers superior
performance across almost all principal evaluation metrics.  
The combination of the asymmetric dual-branch backbone (ADBN), 
the multi-scale feature fusion FPN (MSF-FPN), 
and the ``Droplet'' augmentation strategy 
produces robust feature representations under data-scarce conditions, 
yielding superior pixel-level discrimination and localization performance 
compared to existing methods.

\subsection{Robustness \& Cross-Domain Generalization}
\label{subsec:robustness}

To further evaluate the robustness and cross-domain generalization capability of the CAT framework, we conduct experiments where the model is trained on the KolektorSDD2 (KSDD2) dataset and then tested on three distinct datasets: KolektorSDD1 (KSDD1), Magnetic Tile Defects (MTD), and Rail Surface Defect Detection (RSDD). 
These experiments' results are shown respectively in Table~\ref{tab:kos1_evaluation}, Table~\ref{tab:mtd_evaluation} and Table~\ref{tab:rsdd_evaluation}.
These datasets differ significantly in defect types, surface textures, and imaging conditions from KSDD2, providing a comprehensive test of the model's ability to adapt to unseen domains and avoid overfitting.

\subsubsection{KolektorSDD1 (KSDD1) Evaluation}
KolektorSDD1, as an earlier version of KolektorSDD2 with fewer annotations and incomplete details, was utilized to verify the compatibility of CAT with low-annotation datasets. Experimental results indicated that CAT achieved excellent performance on this dataset: Image-level AUROC = 53.30\%, Pixel-level AUROC = 91.30\%, Per-region Overlap = 62.60\%, $\text{AP}_{\text{det}} = 0.1406$, and $\text{AP}_{\text{loc}} = 0.0021$. 

Compared with other comparative methods, CAT—relying on the general defect features learned from KolektorSDD2—could still effectively identify abnormal regions even under the condition of sparse annotated data. This result demonstrates the strong robustness of CAT against differences between dataset versions.
\begin{table*}[ht]
    \captionsetup{justification=centering, labelsep=newline}

    \centering
    \caption{Evaluation Results on KolektorSDD1 (KSDD1) Dataset}
    \label{tab:kos1_evaluation}
    \scriptsize
    \begin{tabularx}{\textwidth}{l *{6}{>{\centering\arraybackslash}X}}
        \toprule
        Method & Image-level AUROC & Pixel-level AUROC & Per-region Overlap & AP\textsubscript{det} & AP\textsubscript{loc} \\
        \midrule
        SuperSimpleNet & 46.46\% & 78.66\% & 42.80\% & 0.11  & 0.00065 \\
        ReContrast & 52.70\% & 84.30\% & 48.90\% & 0.13  & 0.0014 \\
        MMR & 48.90\% & 90.46\% & 60.09\% & 0.12  & 0.00080 \\
        DFR & 53.00\% & 86.41\% & 36.90\% & \textbf{0.15}  & 0.0014 \\
        PSPNet & 50.00\% & 80.34\% & 0.046\% & 0.13 & 0.00 \\
        CCNet & 50.00\% & 81.50\% & 0.049\% & 0.13 & 0.00 \\
        MobileNet & 50.00\% & 80.06\% & 0.076\% & 0.13 & 0.00 \\
        CAT & \textbf{55.20\%} & \textbf{91.86\%} & \textbf{62.60\%} & 0.14  & \textbf{0.0021} \\
        \bottomrule
    \end{tabularx}
\end{table*}

\subsubsection{Magnetic Tile Defects (MTD) Evaluation}
The MTD dataset focuses on magnetic tile surfaces (non-metallic materials) characterized by complex background textures and ultra-small defects (e.g., pinholes with diameter $<$3px, hairline cracks with width $<$1px), which is utilized to verify CAT's cross-material generalization ability. 

Experimental results demonstrated that CAT achieved promising performance metrics: Image-level AUROC = 55.20\%, Pixel-level AUROC = 89.00\%, Per-region Overlap = 54.00\%, $AP_{\text{det}} = 0.3265$, and $AP_{\text{loc}} = 0.0225$. 

Despite the significant material differences between magnetic tiles and metal surfaces, CAT could still effectively distinguish defects from texture noise. With its pixel-level discrimination and detection accuracy superior to those of existing methods, CAT has verified its strong adaptability to cross-material scenarios.

\begin{table*}[ht]
    \captionsetup{justification=centering, labelsep=newline}

    \centering
    \caption{Evaluation Results on Magnetic Tile Defects (MTD) Dataset}
    \label{tab:mtd_evaluation}
    \scriptsize
    \begin{tabularx}{\textwidth}{l *{6}{>{\centering\arraybackslash}X}}
        \toprule
        Method & Image-level AUROC & Pixel-level AUROC & Per-region Overlap & AP\textsubscript{det} & AP\textsubscript{loc}\\
        \midrule
        SuperSimpleNet & 53.55\% & 82.74\% & 63.30\% & 0.32 & 0.022 \\
        ReContrast & 51.50\% & 85.80\% & 54.50\% & 0.30 & 0.025 \\
        MMR & 55.70\% & 84.03\% & 54.00\% & 0.35 & 0.022 \\
        DFR & 54.27\% & 85.05\% & 48.87\% & 0.33  & 0.023 \\
        PSPNet & 50.00\% & 73.77\% & 0.9920\% & 0.29 & 0.0028 \\ 
        CCNet & 50.00\% & 72.60\% & 1.054\% & 0.29 & 0.0016 \\ 
        MobileNet & 51.00\% & 57.72\% & 1.379\% & 0.29 & 0.0012 \\
        CAT & \textbf{56.00\%} & \textbf{89.00\%} & \textbf{65.20\%} & \textbf{0.36} & \textbf{0.032} \\
        \bottomrule
    \end{tabularx}
\end{table*}

\subsubsection{Rail Surface Defect Detection (RSDD) Evaluation}
The RSDD dataset targets rail surfaces (large-scale scenes) with long thin scratches (50-200px length, 1-3px width) and illumination fluctuations (sunlight overexposure, tunnel underexposure), which is used to verify CAT's cross-scene robustness.

Experimental results showed that CAT achieved Pixel-level AUROC = 89.00\% and Per-region Overlap = 65.20\%, both of which are superior to other comparison methods. Although Image-level AUROC and $AP_{\text{det}}$ were marked as "meaningless (all abnormal samples)" due to data characteristics, the pixel-level and region-level metrics demonstrated that CAT can handle illumination changes and large-scale surface defects, showing excellent cross-scene generalization ability.

\begin{table*}[ht]
    \captionsetup{justification=centering, labelsep=newline}

    \centering
    \caption{Evaluation Results on Rail Surface Defect Detection (RSDD) Dataset}
    \label{tab:rsdd_evaluation}
    \scriptsize
    \begin{tabularx}{\textwidth}{l *{6}{>{\centering\arraybackslash}X}}
        \toprule
        Method & Image-level AUROC & Pixel-level AUROC & Per-region Overlap & AP\textsubscript{det} & AP\textsubscript{loc} \\
        \midrule
        SuperSimpleNet & - & 92.90\% & 32.30\% & - & 0.0023 \\
        ReContrast & - & 91.80\% & 44.00\% & - & 0.0050 \\
        MMR & - & 96.32\% & 43.90\% & - & 0.0038 \\
        DFR & - & 93.40\% & 33.80\% & - & 0.0001 \\
        PSPNet & - & 95.80\% & 1.925\% & - & 0.0000 \\
        CCNet & - & 96.76\% & 2.716\% & - & 0.0000 \\ 
        MobileNet & - & 90.95\% & 16.66\% & - & 0.0041 \\
        CAT & - & \textbf{97.89\%} & \textbf{45.00\%} & - & \textbf{0.0058} \\
        \bottomrule
    \end{tabularx}
\end{table*}

In summary, the CAT framework, when trained on the KSDD2 dataset, exhibits strong robustness and cross-domain generalization ability across KolektorSDD1, MTD, and RSDD datasets. It can adapt to different defect types, surface textures, and imaging conditions, effectively avoiding overfitting to the training dataset and providing a reliable solution for industrial surface defect detection in various scenarios.

\subsection{Failure Cases \& Reasons}
\label{subsec:failure_cases}

Despite the excellent performance of the Contrastive Augmented Transformer (CAT) framework across various metrics, two key limitations have been observed in practical applications and cross-domain deployment.

During the model’s training and inference, to meet the Transformer’s requirement for uniform input size, original non-square images are processed with zero-padding to convert them into square images.
While this approach improves the efficiency and accuracy of feature extraction, it introduces the following potential issues: 
First, feature dilution may occur. In certain specific surface defect detection scenarios (e.g., when the background is gray or white), defect features themselves may appear as pure black. 
After the images undergo black zero-padding, the newly introduced large-area pure black background (I=0) may interfere with or dilute the model in the feature space—especially in shallow feature extraction, which may reduce the model’s sensitivity to defect features. 
Second, transfer learning is affected. When the model is transferred to new scenarios with significantly different background colors or defect-free samples with highly uniform features, the unrealistic edge and color information introduced by zero-padding may reduce the model’s generalization ability.
A potential mitigation strategy involves adopting adaptive padding or content-aware resizing mechanisms to preserve the original spatial distribution and intensity characteristics of surface textures, thereby alleviating the side effects introduced by hard zero-padding.

The second limitation manifests in the conversion process from the anomaly map to the final binary segmentation map, specifically the restriction of the fixed threshold (e.g., 0.5) strategy on the model’s generalization. On the one hand, there exist threshold dependence and transfer problems. In cross-domain transfer experiments (e.g., on KSDD1, MTD, and RSDD datasets), due to the significant differences between the new datasets and KSDD2 in terms of image texture, illumination conditions, and defect contrast, the anomaly scores of defect regions predicted by the model—though higher than those of normal backgrounds—may not reach the fixed 0.5 threshold. This directly leads to a decline in two key metrics: For Image-level AUROC and AP\textsubscript{det}, although images are correctly identified as abnormal (with high scores), the number of false negatives (where images are ultimately judged as normal) increases because precise segmentation cannot be achieved via the 0.5 threshold, lowering the image-level accuracy and detection precision. On the other hand, visualization errors occur. In qualitative visualization results, it can be observed that actual high-confidence defect regions are incorrectly erased due to their scores being lower than 0.5.
This limitation could be alleviated by introducing adaptive or learnable thresholding mechanisms—such as percentile-based dynamic thresholds or local contrast-guided calibration—that adjust segmentation decisions according to dataset statistics or feature-map distributions, thereby improving the model’s robustness to cross-domain score variations.

\begin{figure}[!t]
    \centering
    \includegraphics[width=0.5\textwidth]{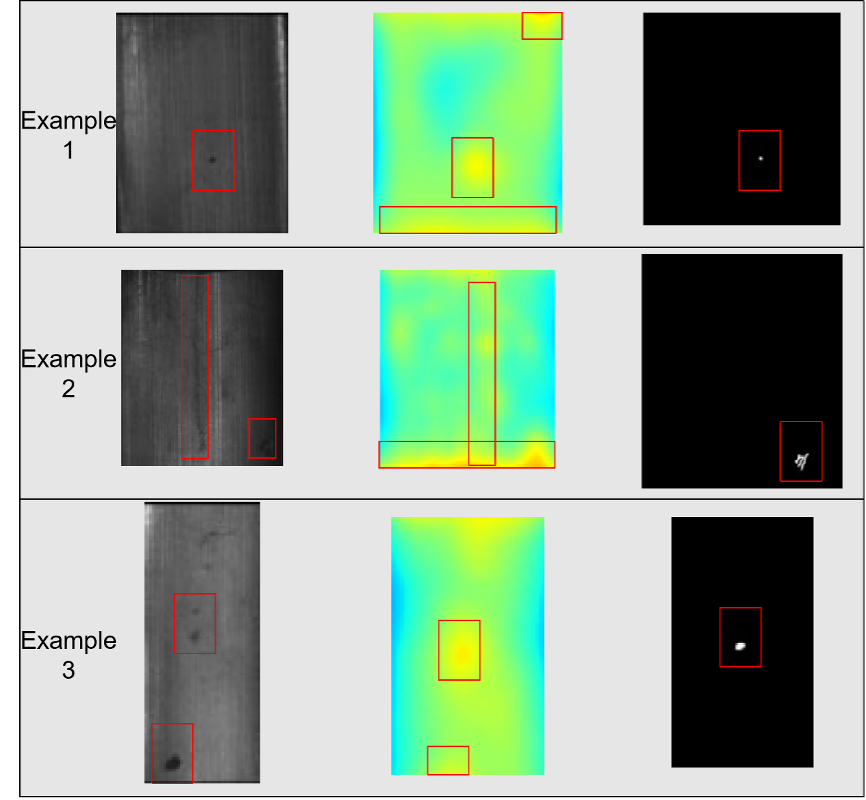}
    \caption{Qualitative visualization of failure cases in the CAT framework, highlighting issues with zero-padding and threshold-based segmentation.}
    \label{fig:failureCase}
\end{figure}
\section{Conclusion}
In recent years, the growing demand for high-precision and high-reliability manufacturing has driven continuous advancements in automated metal surface defect detection. 
Despite remarkable progress, existing approaches still struggle with challenges such as data imbalance, limited defect diversity, and poor generalization across different material types and lighting conditions. 
These issues significantly hinder the deployment of defect detection systems in real-world industrial production lines, where unseen conditions and domain shifts are inevitable.

This paper proposed the Contrastive Augmented Transformer (CAT) framework, 
which effectively addresses the challenges of data scarcity, multi-scale recognition, and insufficient cross-scenario generalization in the field of metal surface defect detection. 
By integrating the hierarchical feature extraction capability of the Swin Transformer, a redesigned Feature Pyramid Network (FPN), 
and a contrastive learning strategy that incorporates Hard Negative Mining, 
CAT significantly enhances the model's accuracy in identifying subtle defect patterns and its discrimination ability in complex backgrounds. 
Experimental results demonstrate that CAT achieves a pixel-level AUROC of 99.54\% on the KolektorSDD2 dataset and exhibits superior cross-domain generalization and 
robustness across a series of unseen datasets (KSDD1, MTD, RSDD), fully validating its substantial potential for application in real-world industrial environments.

However, there are areas for future improvement in this study. 
The current implementation involves a zero-padding operation on the network input, which may restrict the completeness of feature representation and thus impact the model's generalization performance. 
More critically, the fixed threshold (e.g., 0.5) strategy used for converting the anomaly score map to the final binary segmentation map exposes significant threshold dependence and transfer problems when dealing with cross-domain datasets (e.g., those with noticeable texture and illumination differences). 
This directly leads to a decline in image-level detection metrics. To further enhance the practicality and robustness of the CAT framework, future work will focus on developing adaptive or threshold-free defect segmentation methods to eliminate reliance on preset parameters. 
Concurrently, we will explore more optimized data pre-processing and feature fusion mechanisms to ensure the model maintains high accuracy and strong generalization across a wider range of extreme industrial scenarios.

\section*{CRediT authorship contribution statement}
\noindent Yiyao Liu: Literature review, Conceptualization, Methodology, Writing -- original draft. \\
Wenxiao He: Software, Investigation, Validation, Formal analysis, Data curation, Visualization, Writing -- original draft. \\
Liyuan Ren: Supervision, Project administration, Writing -- review \& editing. \\
Huan Wang: Supervision, Writing -- review \& editing.
\section*{Acknowledgements}
This work was supported by the Innovation Fund of Glasgow College, University of Electronic Science and Technology of China. The authors would like to express their sincere gratitude to Prof. Huan Wang for his patient guidance, insightful comments, and generous support throughout this study.
\section*{Declaration of competing interest}
The authors declare that they have no known competing financial interests or personal relationships that could have appeared to influence the work reported in this paper.
\bibliographystyle{elsarticle-num}
\bibliography{new_template/ref}

@INPROCEEDINGS{moco,
  author={He, Kaiming and Fan, Haoqi and Wu, Yuxin and Xie, Saining and Girshick, Ross},
  booktitle={2020 IEEE/CVF Conference on Computer Vision and Pattern Recognition (CVPR)}, 
  title={Momentum Contrast for Unsupervised Visual Representation Learning}, 
  year={2020},
  volume={},
  number={},
  pages={9726-9735},
  doi={10.1109/CVPR42600.2020.00975}}

@article{mocov2,
  title={Improved Baselines with Momentum Contrastive Learning},
  author={Xinlei Chen and Haoqi Fan and Ross B. Girshick and Kaiming He},
  journal={ArXiv},
  year={2020},
  volume={abs/2003.04297},
  url={https://api.semanticscholar.org/CorpusID:212633993}
}

@ARTICLE{Canny,
  author={Canny, John},
  journal={IEEE Transactions on Pattern Analysis and Machine Intelligence}, 
  title={A Computational Approach to Edge Detection}, 
  year={1986},
  volume={PAMI-8},
  number={6},
  pages={679-698},
  doi={10.1109/TPAMI.1986.4767851}}

@InProceedings{U-Net,
author="Ronneberger, Olaf
and Fischer, Philipp
and Brox, Thomas",
editor="Navab, Nassir
and Hornegger, Joachim
and Wells, William M.
and Frangi, Alejandro F.",
title="U-Net: Convolutional Networks for Biomedical Image Segmentation",
booktitle="Medical Image Computing and Computer-Assisted Intervention -- MICCAI 2015",
year="2015",
publisher="Springer International Publishing",
address="Cham",
pages="234--241",
isbn="978-3-319-24574-4"
}

@InProceedings{DeepLabV3+,
author="Chen, Liang-Chieh
and Zhu, Yukun
and Papandreou, George
and Schroff, Florian
and Adam, Hartwig",
editor="Ferrari, Vittorio
and Hebert, Martial
and Sminchisescu, Cristian
and Weiss, Yair",
title="Encoder-Decoder with Atrous Separable Convolution for Semantic Image Segmentation",
booktitle="Computer Vision -- ECCV 2018",
year="2018",
publisher="Springer International Publishing",
address="Cham",
pages="833--851",
isbn="978-3-030-01234-2"
}

@INPROCEEDINGS{resnet,
  author={He, Kaiming and Zhang, Xiangyu and Ren, Shaoqing and Sun, Jian},
  booktitle={2016 IEEE Conference on Computer Vision and Pattern Recognition (CVPR)}, 
  title={Deep Residual Learning for Image Recognition}, 
  year={2016},
  volume={},
  number={},
  pages={770-778},
  doi={10.1109/CVPR.2016.90}}

@article{Automotive,
   title={Fully-Synthetic Training for Visual Quality Inspection in Automotive Production},
   volume={134},
   ISSN={2212-8271},
   url={http://dx.doi.org/10.1016/j.procir.2025.02.205},
   DOI={10.1016/j.procir.2025.02.205},
   journal={Procedia CIRP},
   publisher={Elsevier BV},
   author={Huber, Christoph and Knoll, Dino and Guthe, Michael},
   year={2025},
   pages={777–782} }

@article{Automotive2,
   title={Automated Defect Recognition of Castings Defects Using Neural Networks},
   volume={41},
   ISSN={1573-4862},   url={http://dx.doi.org/10.1007/s10921-021-00842-1},
   DOI={10.1007/s10921-021-00842-1},
   number={1},
   journal={Journal of Nondestructive Evaluation},
   publisher={Springer Science and Business Media LLC},
   author={García Pérez, A. and Gómez Silva, M. J. and de la Escalera Hueso, A.},
   year={2021},
   month=dec 
   }

@article{Automotive3,
author = {Campos, Mario and Martins, Teresa and Ferreira, Manuel and Santos, Cristina},
year = {2008},
month = {06},
pages = {},
title = {Detection of defects in automotive metal components through computer vision},
doi = {10.1109/ISIE.2008.4677037}
}

@Article{aero1,
AUTHOR = {Bounenni, Leith and Arbane, Mohamed and Ibarra-Castanedo, Clemente and Yaddaden, Yacine and Unnikrishnakurup, Sreedhar and Yong, Andrew Ngo Chun and Maldague, Xavier},
TITLE = {Advanced Defect Detection on Curved Aeronautical Surfaces Through Infrared Imaging and Deep Learning},
JOURNAL = {NDT},
VOLUME = {2},
YEAR = {2024},
NUMBER = {4},
PAGES = {519--531},
URL = {https://www.mdpi.com/2813-477X/2/4/32},
ISSN = {2813-477X},
DOI = {10.3390/ndt2040032}
}

@INPROCEEDINGS{aero2,
  author={Agarwal, Arpit and Ajith, Abhiroop and Wen, Chengtao and Stryzheus, Veniamin and Miller, Brian and Chen, Matthew and Johnson, Micah K. and Susa Rincon, Jose Luis and Rosca, Justinian and Yuan, Wenzhen},
  booktitle={2023 IEEE/RSJ International Conference on Intelligent Robots and Systems (IROS)}, 
  title={Robotic Defect Inspection with Visual and Tactile Perception for Large-Scale Components}, 
  year={2023},
  volume={},
  number={},
  pages={10110-10116},
  doi={10.1109/IROS55552.2023.10341590}}

@Article{pm1,
AUTHOR = {Zeng, Li and Wan, Feng and Zhang, Baiyun and Zhu, Xu},
TITLE = {Automated Visual Inspection for Precise Defect Detection and Classification in CBN Inserts},
JOURNAL = {Sensors},
VOLUME = {24},
YEAR = {2024},
NUMBER = {23},
ARTICLE-NUMBER = {7824},
URL = {https://www.mdpi.com/1424-8220/24/23/7824},
PubMedID = {39686361},
ISSN = {1424-8220},
DOI = {10.3390/s24237824}
}

@Article{pm2,
AUTHOR = {Xu, Xiujin and Zhang, Gengming and Zheng, Wenhe and Zhao, Anbang and Zhong, Yi and Wang, Hongjun},
TITLE = {High-Precision Detection Algorithm for Metal Workpiece Defects Based on Deep Learning},
JOURNAL = {Machines},
VOLUME = {11},
YEAR = {2023},
NUMBER = {8},
ARTICLE-NUMBER = {834},
URL = {https://www.mdpi.com/2075-1702/11/8/834},
ISSN = {2075-1702},
DOI = {10.3390/machines11080834}
}

@article{mmr,
title = {Industrial anomaly detection with domain shift: A real-world dataset and masked multi-scale reconstruction},
journal = {Computers in Industry},
volume = {151},
pages = {103990},
year = {2023},
issn = {0166-3615},
doi = {https://doi.org/10.1016/j.compind.2023.103990},
url = {https://www.sciencedirect.com/science/article/pii/S0166361523001409},
author = {Zilong Zhang and Zhibin Zhao and Xingwu Zhang and Chuang Sun and Xuefeng Chen}
}

@INPROCEEDINGS{MTD,
  author={Huang, Yibin and Qiu, Congying and Guo, Yue and Wang, Xiaonan and Yuan, Kui},
  booktitle={2018 IEEE 14th International Conference on Automation Science and Engineering (CASE)}, 
  title={Surface Defect Saliency of Magnetic Tile}, 
  year={2018},
  volume={},
  number={},
  pages={612-617},
  doi={10.1109/COASE.2018.8560423}}

@INPROCEEDINGS{CutPaste,
  author={Li, Chun-Liang and Sohn, Kihyuk and Yoon, Jinsung and Pfister, Tomas},
  booktitle={2021 IEEE/CVF Conference on Computer Vision and Pattern Recognition (CVPR)}, 
  title={CutPaste: Self-Supervised Learning for Anomaly Detection and Localization}, 
  year={2021},
  volume={},
  number={},
  pages={9659-9669},
  doi={10.1109/CVPR46437.2021.00954}}

@InProceedings{NSA,
author="Schl{\"u}ter, Hannah M.
and Tan, Jeremy
and Hou, Benjamin
and Kainz, Bernhard",
editor="Avidan, Shai
and Brostow, Gabriel
and Ciss{\'e}, Moustapha
and Farinella, Giovanni Maria
and Hassner, Tal",
title="Natural Synthetic Anomalies for Self-supervised Anomaly Detection and Localization",
booktitle="Computer Vision -- ECCV 2022",
year="2022",
publisher="Springer Nature Switzerland",
address="Cham",
pages="474--489",
isbn="978-3-031-19821-2"
}

@article{MAEDAY,
title = {MAEDAY: MAE for few- and zero-shot AnomalY-Detection},
journal = {Computer Vision and Image Understanding},
volume = {241},
pages = {103958},
year = {2024},
issn = {1077-3142},
doi = {https://doi.org/10.1016/j.cviu.2024.103958},
url = {https://www.sciencedirect.com/science/article/pii/S1077314224000390},
author = {Eli Schwartz and Assaf Arbelle and Leonid Karlinsky and Sivan Harary and Florian Scheidegger and Sivan Doveh and Raja Giryes}
}

@article{KOS2,
  author = {Bo{\v{z}}i{\v{c}}, Jakob and Tabernik, Domen and 
  Sko{\v{c}}aj, Danijel},
  journal = {Computers in Industry},
  title = {{Mixed supervision for surface-defect detection:
from weakly to fully supervised learning}},
  year = {2021}
}

@inproceedings{Gabor,
author = {Liu, Bin and Wu, Shengjin and Zou, Shifang},
year = {2010},
month = {07},
pages = {2213 - 2216},
title = {Automatic detection technology of surface defects on plastic products based on machine vision},
journal = {2010 International Conference on Mechanic Automation and Control Engineering, MACE2010},
doi = {10.1109/MACE.2010.5536470}
}

@InProceedings{Texture,
author="V., Asha
and N.U., Bhajantri
and P., Nagabhushan",
editor="Venugopal, K. R.
and Patnaik, L. M.",
title="Automatic Detection of Texture Defects Using Texture-Periodicity and Gabor Wavelets",
booktitle="Computer Networks and Intelligent Computing",
year="2011",
publisher="Springer Berlin Heidelberg",
address="Berlin, Heidelberg",
pages="548--553",
isbn="978-3-642-22786-8"
}

@article{swin2,
author = {Gao, Linfeng and Zhang, Jianxun and Yang, Changhui and Zhou, Yuechuan},
title = {Cas-VSwin transformer: A variant swin transformer for surface-defect detection},
year = {2022},
issue_date = {Sep 2022},
publisher = {Elsevier Science Publishers B. V.},
address = {NLD},
volume = {140},
number = {C},
issn = {0166-3615},
url = {https://doi.org/10.1016/j.compind.2022.103689},
doi = {10.1016/j.compind.2022.103689},
journal = {Comput. Ind.},
month = sep,
numpages = {10}
}

@InProceedings{ssn1,
  author={Rolih, Bla{\v{z}} and Fu{\v{c}}ka, Matic and Sko{\v{c}}aj, Danijel},
  booktitle={International Conference on Pattern Recognition}, 
  title={{S}uper{S}imple{N}et: {U}nifying {U}nsupervised and {S}upervised {L}earning for {F}ast and {R}eliable {S}urface {D}efect {D}etection},
  year={2024}
}

@article{ssn2,
  author={Rolih, Bla{\v{z}} and Fu{\v{c}}ka, Matic and Sko{\v{c}}aj, Danijel},
  journal={Journal of Intelligent Manufacturing}, 
  title={No Label Left Behind: A Unified Surface Defect Detection Model for all Supervision Regimes},
  year={2025}
}

@inproceedings{recontrast,
 author = {Guo, Jia and Lu, Shuai and Jia, Lize and Zhang, Weihang and Li, Huiqi},
 booktitle = {Advances in Neural Information Processing Systems},
 pages = {10721--10740},
 title = {ReContrast: Domain-Specific Anomaly Detection via Contrastive Reconstruction},
 volume = {36},
 year = {2023}
}

@article{dfr,
title = {Unsupervised anomaly segmentation via deep feature reconstruction},
journal = {Neurocomputing},
volume = {424},
pages = {9-22},
year = {2021},
issn = {0925-2312},
doi = {https://doi.org/10.1016/j.neucom.2020.11.018},
url = {https://www.sciencedirect.com/science/article/pii/S0925231220317951},
author = {Yong Shi and Jie Yang and Zhiquan Qi}
}

@INPROCEEDINGS{pspnet,
  author={Zhao, Hengshuang and Shi, Jianping and Qi, Xiaojuan and Wang, Xiaogang and Jia, Jiaya},
  booktitle={2017 IEEE Conference on Computer Vision and Pattern Recognition (CVPR)}, 
  title={Pyramid Scene Parsing Network}, 
  year={2017},
  volume={},
  number={},
  pages={6230-6239},
  doi={10.1109/CVPR.2017.660}}

@INPROCEEDINGS{ccnet,
  author={Huang, Zilong and Wang, Xinggang and Huang, Lichao and Huang, Chang and Wei, Yunchao and Liu, Wenyu},
  booktitle={2019 IEEE/CVF International Conference on Computer Vision (ICCV)}, 
  title={CCNet: Criss-Cross Attention for Semantic Segmentation}, 
  year={2019},
  volume={},
  number={},
  pages={603-612},
  doi={10.1109/ICCV.2019.00069}}

@article{mobilenet,
  title={MobileNets: Efficient Convolutional Neural Networks for Mobile Vision Applications},
  author={Andrew G. Howard and Menglong Zhu and Bo Chen and Dmitry Kalenichenko and Weijun Wang and Tobias Weyand and Marco Andreetto and Hartwig Adam},
  journal={ArXiv},
  year={2017},
  volume={abs/1704.04861},
  url={https://api.semanticscholar.org/CorpusID:12670695}
}

@article{DeepLabV3,
  title={Rethinking Atrous Convolution for Semantic Image Segmentation},
  author={Liang-Chieh Chen and George Papandreou and Florian Schroff and Hartwig Adam},
  journal={ArXiv},
  year={2017},
  volume={abs/1706.05587},
  url={https://api.semanticscholar.org/CorpusID:22655199}
}

@article{R1-LRD,
author = {Hou, Yingqiang and Zhang, Xindong},
year = {2025},
month = {09},
pages = {122685},
title = {A lightweight real-time detection transformer model for surface defect detection systems},
volume = {725},
journal = {Information Sciences},
doi = {10.1016/j.ins.2025.122685}
}

@article{R3-SDDC,
author = {Zhu, Wei and Zhang, Hui and Zhang, Chao and Zhu, Xiaoyang and Guan, Zhen and Jia, Jiale},
year = {2023},
month = {08},
pages = {102061},
title = {Surface defect detection and classification of steel using an efficient Swin Transformer},
volume = {57},
journal = {Advanced Engineering Informatics},
doi = {10.1016/j.aei.2023.102061}
}

@article{R2-ACT,
author = {Huang, Xiaohua and Li, Yang and Bao, Yongqiang and Zheng, Wenming},
year = {2024},
month = {01},
pages = {1-1},
title = {Adaptive Cross Transformer With Contrastive Learning for Surface Defect Detection},
volume = {PP},
journal = {IEEE Transactions on Instrumentation and Measurement},
doi = {10.1109/TIM.2024.3470998}
}

@article{add_1,
title = {DD-DETR: A dual-decoder DETR with information interaction and competitive learning for blade surface defect detection},
journal = {Advanced Engineering Informatics},
volume = {71},
pages = {104234},
year = {2026},
issn = {1474-0346},
doi = {https://doi.org/10.1016/j.aei.2025.104234},
url = {https://www.sciencedirect.com/science/article/pii/S1474034625011279},
author = {Xiongfeng Shi and Lai Zou and Kefei Qian and Xin Liu},
}

@article{add_2,
title = {Simple and effective Frequency-aware Image Restoration for industrial visual anomaly detection},
journal = {Advanced Engineering Informatics},
volume = {64},
pages = {103064},
year = {2025},
issn = {1474-0346},
doi = {https://doi.org/10.1016/j.aei.2024.103064},
url = {https://www.sciencedirect.com/science/article/pii/S1474034624007158},
author = {Tongkun Liu and Bing Li and Xiao Du and Bingke Jiang and Leqi Geng and Feiyang Wang and Zhuo Zhao}
}

@article{add_3,
title = {AI-enabled defect detection in industrial products: A comprehensive survey, key insights and future research challenges},
journal = {Advanced Engineering Informatics},
volume = {69},
pages = {104067},
year = {2026},
issn = {1474-0346},
doi = {https://doi.org/10.1016/j.aei.2025.104067},
url = {https://www.sciencedirect.com/science/article/pii/S1474034625009607},
author = {Lutfun Nahar and Mohammad Awrangjeb and Md. Saiful Islam}
}
\end{document}